# SadaqahStories: A Verified English-Arabic Corpus of Charity Stories for Children, with an Audit of Errors in AI-Drafted Religious Narratives

Wajdi Zaghouani, Spencer Striker
Northwestern University in Qatar, Doha, Qatar
{wajdi.zaghouani, spencer.striker}@northwestern.edu

**Abstract**

We present SadaqahStories, a parallel English-Arabic corpus of 41 illustrated four-page stories about charitable giving in Islam (Sadaqah, Zakat and Waqf), written for readers aged 7 to 18. The stories were first drafted with large language model (LLM) assistance, then audited against hadith collections, Qur'an references and historical sources, and finally rewritten for children and signed off by an Islamic scholar. Every story carries an evidence label and links to its sources. We release all three versions of each story (AI-assisted draft, audited text, final child-oriented text) together with the correction log, so the corpus can be used to study how AI drafts of religious and historical content go wrong. The audit found substantive errors in 39 of the 41 drafts. The most frequent were unsupported specific details (dates, figures, prices, names), invented quotations, institutions or anecdotes, and conflation of separate events or people; six drafts contained errors in the wording, grading or ruling of a religious text. Rewriting cut the estimated Flesch-Kincaid grade of the English text from 14.0 to 6.1. We also release an 81-entry charity lexicon covering Modern Standard Arabic and Qatari Arabic, and describe the free, account-free website that serves the corpus.

## 1. Introduction

Children in Qatar hear about charity at home, in school and in the many donation campaigns run by local charities. Material that teaches the history and meaning of Islamic giving to children is scarce in digital form, and it is rarely available in both Arabic and English with its sources shown. Generative models make it cheap to produce such stories. They also make it cheap to produce stories that sound right and are wrong: a hadith quoted with the wrong wording, a hospital given an invented founding deed, or a famine relief effort credited with a precise number of camels that no source records. For religious content aimed at children these errors are not a small matter.

This paper describes a corpus built to deal with that problem, and what we learned while building it. Our starting point was a set of 62 bilingual story drafts produced with LLM assistance for an educational website funded by the Qatar Research, Development and Innovation Council and the Regulatory Authority for Charitable Activities (RACA). We kept 41 of them, corrected each against its sources, rewrote them for young readers and had the final text signed off by an Islamic scholar. Along the way we logged every correction.

Our contributions are:

- a parallel English-Arabic corpus of 41 four-page children's stories (328 page-level passages) on Islamic charity, with evidence labels and 58 source links;
- the three versions of every story (draft, audited, final) and a per-story correction log, coded with a simple error typology;
- an analysis of the errors found in the AI-assisted drafts and of how the text changed between versions;
- an 81-entry lexicon of charitable terms in Modern Standard Arabic (MSA) and Qatari Arabic, normalized for release.

The corpus is small. Its value lies in the verification and in the aligned versions, not in its size.

## 2. Related Work

**Arabic reading resources for children.** Educational Arabic corpora are few. Al Khalil et al. (2018) built a leveled reading corpus from UAE curriculum textbooks and fiction, and the BAREC corpus annotates over a million words for sentence-level readability across 19 levels (Elmadani et al., 2025; Habash et al., 2025). These resources describe how hard a text is. They do not address whether a text is true, which is the central issue for historical and religious stories.

**Religious text and LLM errors.** Work on the Qur'an and hadith in NLP has focused on question answering over the canonical texts (Malhas et al., 2022). IslamicEval 2025 targeted LLM hallucination in Islamic content directly, asking systems to detect, validate and correct Qur'anic verses and hadith quoted in model output (Mubarak et al., 2025). Our audit sees the same problem from the side of content production. Misquoted scripture is one error type among several; most errors we found sit in the historical narrative around the scripture. General surveys treat these errors as hallucination (Ji et al., 2023), and work on LLMs in education has warned about them for some time (Kasneci et al., 2023).

**Gulf Arabic and dialect lexicons.** Gulf Arabic has fewer resources than Egyptian or Levantine Arabic. The Gumar corpus collected 110 million words of Gulf Arabic forum novels (Khalifa et al., 2016), and MADAR provides a parallel dialect corpus and lexicon across 25 cities (Bouamor et al., 2018). Our lexicon is tiny by comparison, but it covers a domain (charity, mutual aid, and the formulas used to give and receive thanks) that general dialect lexicons do not target.

**Documentation.** We follow the practice of documenting how a resource was made and what it should not be used for (Bender and Friedman, 2018; Gebru et al., 2021). The correction log is part of that documentation.

# 3. Building the Corpus

| Stage | What was done | Stories |
|---|---|---|
| V0 | LLM-assisted bilingual drafts, four pages each | 62 |
| Selection | Near-duplicates and drafts without usable sources removed | 41 |
| V1 | Editorial audit against sources; corrections logged; evidence labels and links added | 41 |
| V2 | Rewritten for children in both languages; no facts added | 41 |
| Sign-off | Islamic scholar reviewed the final English and Arabic text | 41 |

Table 1: Corpus construction stages.

Table 1 summarizes the process.

**Drafting (V0).** The drafts were produced with LLM assistance in English and Arabic, each as four short pages with one illustration per page. The 62 drafts covered five themes: the Prophetic era and the Companions, welfare under the early caliphs, endowed institutions such as hospitals, schools and libraries, the categories of Zakat recipients, and present-day work by Qatari charities. Illustrations were generated with Adobe Firefly and show ordinary people and settings; no prophet is depicted.

**Selection.** We removed near-duplicate stories, keeping one version of each episode, and drafts for which we could not find a usable source. We also removed stories likely to be contested by religious authorities in Qatar. This left 41 stories.

**Audit (V1).** Each story was checked against the sources available for its claims: hadith collections (with the grading given in the cited edition on sunnah.com), Qur'an references, historical and institutional pages, and the published reports of the charities named. When a claim had no support it was removed or reworded as uncertain. Each change was entered in a log with a one-line description. The audit also assigned an evidence label (Section 4.2) and attached at least one source link to every story.

**Child-oriented rewrite (V2).** The audited text was rewritten in a simpler voice for readers of roughly 9 to 12 in both languages. The rule was that no fact, number, name or quotation could be added. Automated checks compared numbers, citations and quotation marks between V1 and V2, and the English text was read in full against V1. Where the audit had left a claim uncertain, V2 says so in plain words, for example "We cannot be sure of these details."

**Scholarly sign-off.** An Islamic scholar reviewed the final English and Arabic text of all 41 stories, including every Qur'anic and hadith reference, before the stories were published.

# 4. Corpus Description

## 4.1. Size and readability

| | EN V0 | EN V1 | EN V2 | AR V0 | AR V1 | AR V2 |
|---|---|---|---|---|---|---|
| Tokens | 7,277 | 4,534 | 6,304 | 6,585 | 3,439 | 4,696 |
| Types | 2,194 | 1,445 | 1,284 | 3,352 | 1,915 | 2,315 |
| Tokens/page | 44.4 | 27.6 | 38.4 | 40.2 | 21.0 | 28.6 |
| Sentences | 348 | 280 | 461 | 328 | 213 | 377 |
| Tokens/sent. | 20.9 | 16.2 | 13.7 | 20.1 | 16.1 | 12.5 |
| Quotations | 20 | 4 | 2 | 18 | 4 | 3 |
| FK grade | 14.0 | 10.3 | 6.1 | – | – | – |

Table 2: Statistics for the 41 published stories (164 pages per language) across versions. Tokens are whitespace-separated words; quotations are counted from paired quotation marks; FK is the Flesch-Kincaid grade, with syllables estimated by hyphenation.

Table 2 gives statistics for the 41 stories at each stage. Each version has 164 pages per language. The audit shortened the text by about 38% in English and 48% in Arabic, mostly by deleting unsupported detail. The rewrite then lengthened it again, because explaining a term for a child ("A caliph was the leader of the Muslim community") takes words, and because uncertainty now has to be stated rather than left out. Sentences in V2 are a third shorter than in V0 in both languages. The English Flesch-Kincaid grade (Kincaid et al., 1975) fell from 14.0 to 6.1; per story it ranges from 4.4 to 8.9 in V2, against 8.7 to 19.4 in V0. FK is a rough measure and we use it only to show the direction of change. We did not compute an Arabic readability score; a BAREC-based model (Elmadani et al., 2025) would be the natural tool for future work.

Direct quotations dropped from 20 to 2 in English. Many V0 quotations were speech attributed to historical figures with no source. In V2, 17 of the 41 English stories contain at least one explicit statement of uncertainty (29 matches of a small set of hedging patterns), against 2 matches in the same stories in V0.

## 4.2. Themes and evidence labels

| Theme | N | Had. | Doc. | Trad. | Ov. | Oth. |
|---|---|---|---|---|---|---|
| Prophet and Companions | 14 | 10 | 0 | 3 | 0 | 1 |
| Early caliphates | 4 | 0 | 0 | 4 | 0 | 0 |
| Endowed institutions | 12 | 1 | 5 | 1 | 5 | 0 |
| Zakat categories | 7 | 0 | 1 | 0 | 0 | 6 |
| Contemporary Qatar | 4 | 0 | 4 | 0 | 0 | 0 |
| Total | 41 | 11 | 10 | 8 | 5 | 7 |

Table 3: Stories by theme and evidence label. Had.: authenticated hadith cited; Doc.: documented historical or organizational source; Trad.: historical tradition; Ov.: general

historical overview; Oth.: general guidance (3), mixed Qur'an, hadith or fiqh sources (3), educational example (1).

Each story has one evidence label that tells the reader what kind of support it has (Table 3). The labels are a judgment about the core of the story, not about every sentence. “Authenticated hadith cited” means the central event is reported in a hadith graded authentic in the cited edition; the details around it may still be narrative. “Historical tradition” covers well-known accounts from biography and history (for example Umar's relief work in the Year of Ashes) that are not narrated in the canonical hadith collections. “General historical overview” is used for institutions that existed as a type, such as caravanserais or bathhouses held as waqf, where the story does not claim a specific named building. The label distribution follows the themes closely: stories about the Prophet and the Companions rest on hadith, stories about Qatari charities rest on their published reports, and stories about the Zakat categories rest on Qur'an 9:60 and juristic guidance.

The 41 stories carry 58 source links: 27 to sunnah.com, 17 to quran.com, and 14 to historical, museum, government and charity pages. Every link uses HTTPS.

### 4.3. Format

Each story is a JSON record with an identifier, theme, evidence label, sources, English and Arabic titles, four aligned page pairs, the correction log entry and the error codes from Section 5. A second file holds the six versions of each story (V0, V1, V2 in each language) aligned by page. Arabic text keeps the diacritics used in the original (422 diacritic marks in V2), and the honorific ligature for the Prophet (U+FDFA) is kept as written.

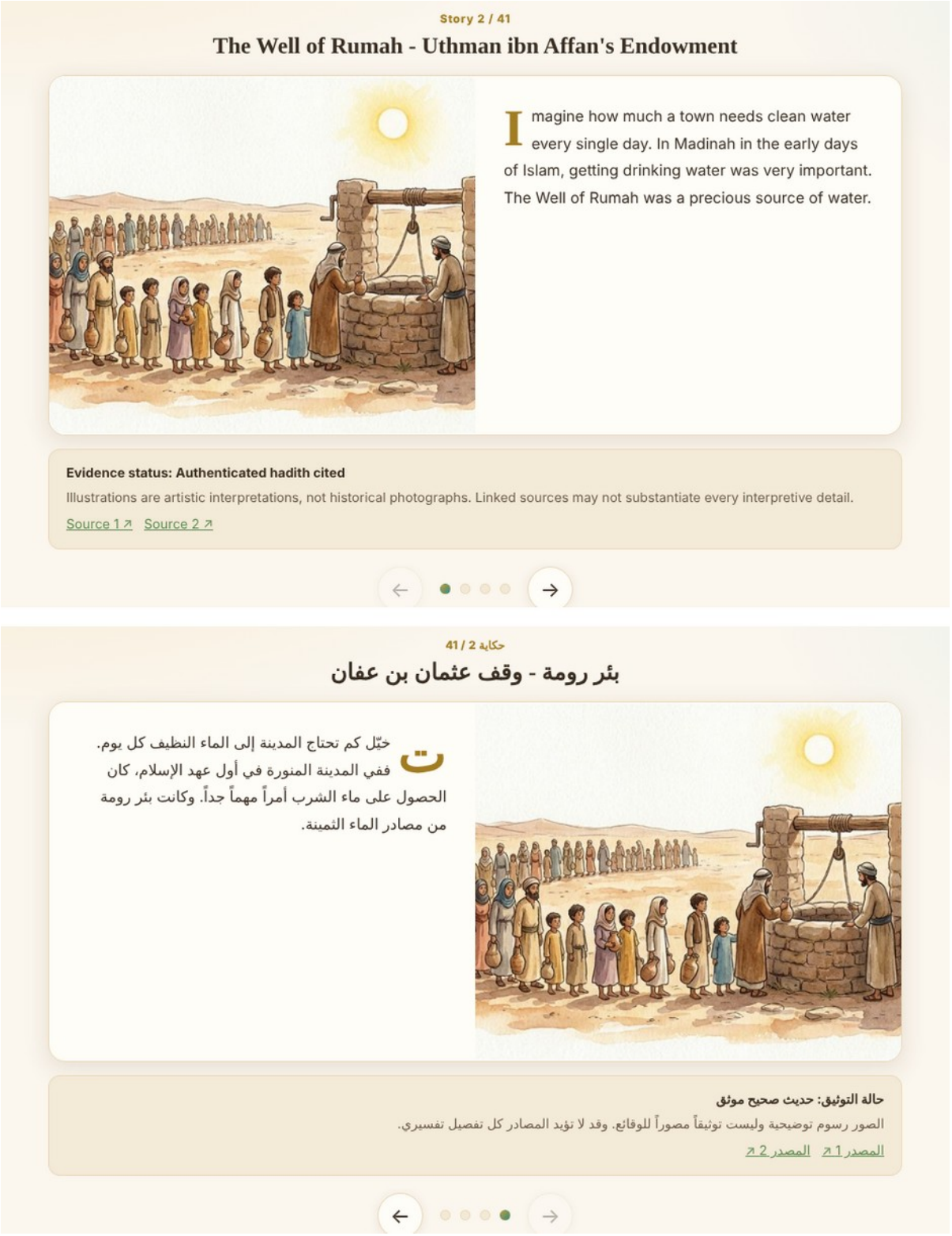

Figure 1: The first page of story 02 (the Well of Rumah) in English (top) and Arabic (bottom) on the website. The panel under each page shows the evidence label, a note that the images are artistic interpretations, and the source links.

## 5. What the Audit Found

We coded the 41 log entries with the categories in Table 4. An entry can have more than one code. The coding was done by one person after the audit, from the log and the V0/V1 text; we did not measure agreement, so the counts should be read as indicative.

| Error type (stories) | Example from the log |
|---|---|
| Unsupported specific detail (16) | Story 05 (Year of Ashes): “over 4,000 camels”, meals “twice daily”, relief for “nine months” removed; sources give different figures or none. |
| Fabricated entity, quotation or anecdote (9) | Story 11: a large Damascus orphanage teaching carpentry and farming to “hundreds of orphans”, not confirmed by early sources; replaced by the reported help for people with leprosy and disabilities. |
| Conflation or misattribution (9) | Story 45: founder of the Salihiyya madrasa confused with the patron of the mausoleum; story 19: two migrations to Abyssinia merged. |
| Overstated certainty or generalization (8) | Story 24: all caravanserais described as offering free lodging; story 37: waqf bathhouses described as free to all. |
| Religious-text error (6) | Story 33: misquoted Qur'an 24:32; story 02: Bukhari 2778 rewards whoever digs the Well of Rumah, the draft said purchases. |
| Historical, arithmetic or anachronism error (3) | Story 04: hospital said to have run “over 400 years” from 981 to 1258 (277 years); story 10: modern hospital departments. |
| Translation error (1) | Story 28 (Arabic): the Prophet urges *sadaqah* in Sahih Muslim 1017a; the draft wrote الزكاة, the obligatory alms. |
| Source anchoring only (7) | Story 17: no error found; narrative tied explicitly to Sahih Muslim 2984a. |

Table 4: Error typology for the correction log of the 41 published stories, with the number of stories in which each type occurs.

Thirty-nine of the 41 drafts had at least one substantive error. Only stories 17 and 60 needed nothing more than a clearer link to their source. Across the 39 stories we recorded 52 substantive codes.

**The narrative around the scripture.** Six stories had an error in a Qur'anic or hadith text or in a ruling drawn from it: a misquoted verse (story 33), hadith wording that did not match the cited collection (stories 02 and 16), a saying attributed to the Prophet with no reference (story 52), a Zakat al-Fitr story that misstated the obligation and passed over the disagreement on paying it in cash (story 18), and the Arabic rendering discussed below (story 28). Outside the 41 stories, one of the model texts on the Community Wall quoted a hadith graded weak as if it were authentic. These are the errors that IslamicEval targets (Mubarak et al., 2025). They were not the most common type. Most errors sat in the historical frame: numbers, dates, prices, staff counts, named campaigns and direct speech that made a story vivid and had no source. A story about Abdurrahman bin Awf said he left half his fortune to charity; the cited hadith does not say so. A story about the Adudi hospital in Baghdad listed four departments and 24 physicians; we could not confirm these figures from the sources at hand.

**Themes differ.** In stories about the Prophet and the Companions, 5 of the 14 had a religious-text or translation error. In stories about institutions and the Zakat categories, invented or generalized institutions dominated: an Ottoman tree-protection waqf with a permit system, a unified fund for travelers, guild finance schemes. The four stories about Qatari charities had unverified campaign names, beneficiary counts and clinical outcomes; we cut them back to what the charities' own published reports state.

**Arabic needs its own check.** The Arabic drafts were not a simple translation of the English ones, and they had their own errors. The clearest case is story 28, where the Arabic draft turned voluntary charity into Zakat, the obligatory alms (Table 4). Its English counterpart was correct. Checking only one language would have missed this.

**What correction costs.** Correcting a draft usually meant deleting material. V1 is 38% shorter than V0 in English. Children's stories need concrete detail, and much of the deleted detail was what made the drafts read well. In V2 we put some of that interest back without new facts: by explaining terms, and by telling the reader openly that "some stories" say more than the sources support. We think this is also a useful lesson for children about how history is known.

## 6. The Charity Lexicon

| Group | N | Example |
|---|---|---|
| Core concepts | 20 | صَدَقَة جَارِيَة *ṣadaqah jāriyah* |
| Charitable sector | 22 | كَفَالَة يَتِيم *kafālat yatīm* |
| Gulf vocabulary | 20 | فَزْعَة *faz*ʿ*ah* |
| Formulaic expressions | 9 | لا شُكْر عَلى واجِب *lā shukr* ʿ*alā wājib* |
| Qatari institutions | 6 | صُنْدُوق الزَّكَاة *ṣundūq al-zakāh* |
| Campaign slogans | 4 | تْبَرَّع الحِين *tbarra*ʿ *al-ḥīn* |
| Total | 81 | |

Table 5: Lexicon groups with one example each.

The lexicon has 81 entries, each with the Arabic form (fully diacritized), a transliteration, and a usage note in English and in Arabic (Table 5). About half the entries are MSA terms from Qur'anic, juristic and institutional use, such as زَكَاة، وَقْف، كَفَّارَة. The rest are Gulf and Qatari: dialect pronunciations such as *ṣadgah* and *fagīr* with /q/ realized as [g], words for collective help such as فَزْعَة and عَوْنَة, the formulas used to thank a donor and to refuse thanks, the names of Qatari charitable bodies, and slogans seen on donation posters. These are words children in Qatar hear, and several of them do not appear in MSA dictionaries with this sense.

The lexicon was first prepared in a word processor, and its extracted text had problems that are common in Arabic resources: 74 of 81 Arabic forms used presentation-form code points rather than standard Arabic letters, some letters came out as their Persian equivalents (*yeh* U+06CC, *heh doachashmee* U+06BE) after normalization, Latin ligatures appeared in the English notes, and six transliterations had lost word spaces ("al-hayʾahal-khayriyya h"). The released file applies NFKC normalization, maps the Persian code points back to Arabic *yeh* and *heh*, and repairs the spacing. Without this step, string lookup against standard Arabic text fails. Tools such as CAMeL Tools (Obeid et al., 2020) can then be applied for morphological analysis.

## 7. The Platform

The corpus is served by a free, bilingual website at https://sadaqahjourney.com (Figure 1). Besides the stories it includes an impact simulator for nine kinds of charitable projects over 1, 5, 10 and 20 years, nine games for three age bands (7 to 10, 11 to 14, 15 to 18), a guided page where children can write their own stories of giving, and the lexicon. The site is static. It has no accounts, cookies or analytics, and progress and children's own stories are stored only in the browser. We chose this design so that no personal data from minors is collected, in line with Qatar's personal data protection law, and so that the site can run with no hosting cost after the funding period. No LLM runs on the site: everything a child reads was written and checked in advance.

## 8. Uses of the Resource

We see three uses. First, the stories are reading material for schools and families, with sources a teacher can open. Second, the aligned V0-V1-V2 versions with the correction log form a small test set for detecting and correcting errors in generated religious and historical narrative, beyond the verse and hadith spans covered by existing tasks. Third, the V1-V2 pairs are a parallel set for simplification of informational text for children in English and Arabic, where the constraint is that facts must not change.

## 9. Conclusion

SadaqahStories is a small, carefully checked bilingual corpus of children's stories about Islamic charity, released with its drafts and corrections. The audit shows that LLM-assisted drafts of this kind need line-by-line checking: 39 of 41 contained at least one substantive error, and most of the errors were in historical detail rather than in scripture. Next steps are to measure how children read and use the stories in classroom pilots, to score the Arabic text with an Arabic readability model, and to add stories from the remaining drafts as they pass review.

## Limitations

The corpus is small (41 stories, about 6,300 English and 4,700 Arabic tokens in the final version) and covers one domain. The error coding was done by one coder without an agreement study, and the typology was built for this log. The audit relied on sources available online; some historical claims were marked uncertain because we could not reach primary sources, not because they are false. The evidence labels describe the core of each story only. Readability was estimated for English with a formula that was not designed for children's text and was not computed for Arabic. We have not yet evaluated the stories with children, and we make no claim about learning outcomes. Finally, we do not report which LLM produced each V0 draft, so the error rates should not be read as a property of any one model.

## Ethics Statement

The stories concern religious figures and practices. All final text was reviewed by an Islamic scholar, and illustrations do not depict prophets. Uncertain claims are stated as uncertain to the child reader rather than presented as fact. The V0 drafts contain errors, including misquoted scripture; they are released for research only, are marked as unverified, and should not be shown to children or used as a source of religious knowledge. The website collects no personal data. Stories about Qatari charities report only what those organizations have published.

## Data Availability

The corpus (final stories with metadata, aligned V0-V1-V2 versions, correction log and error codes) and the normalized lexicon will be released under CC BY-NC 4.0 with the website source code at https://github.com/wajdiz/sadaqahjourney.

## Acknowledgements

This work was funded by the Qatar Research, Development and Innovation Council (QRDI) and the Regulatory Authority for Charitable Activities (RACA) under grant MCSC02-0217-250013. The statements made herein are solely the responsibility of the authors. We thank the Islamic scholar who reviewed the stories.

## Bibliographical References

Muhamed Al Khalil, Hind Saddiki, Nizar Habash, and Latifa Alfalasi. 2018. A leveled reading corpus of Modern Standard Arabic. In Proceedings of the Eleventh International Conference on Language Resources and Evaluation (LREC 2018), Miyazaki, Japan. European Language Resources Association (ELRA).

Emily M. Bender and Batya Friedman. 2018. Data statements for natural language processing: Toward mitigating system bias and enabling better science. Transactions of the Association for Computational Linguistics, 6:587–604.

Houda Bouamor, Nizar Habash, Mohammad Salameh, Wajdi Zaghouani, Owen Rambow, Dana Abdulrahim, Ossama Obeid, Salam Khalifa, Fadhl Eryani, Alexander Erdmann, and Kemal Oflazer. 2018. The MADAR Arabic dialect corpus and lexicon. In Proceedings of the Eleventh International Conference on Language Resources and Evaluation (LREC 2018), Miyazaki, Japan. European Language Resources Association (ELRA).

Khalid N. Elmadani, Nizar Habash, and Hanada Taha-Thomure. 2025. A large and balanced corpus for fine-grained Arabic readability assessment. In Findings of the Association for Computational Linguistics: ACL 2025, pages 16376–16400, Vienna, Austria. Association for Computational Linguistics.

Timnit Gebru, Jamie Morgenstern, Briana Vecchione, Jennifer Wortman Vaughan, Hanna Wallach, Hal Daumé III, and Kate Crawford. 2021. Datasheets for datasets. Communications of the ACM, 64(12):86–92.

Nizar Habash, Hanada Taha-Thomure, Khalid N. Elmadani, Zeina Zeino, and Abdallah Abushmaes. 2025. Guidelines for fine-grained sentence-level Arabic readability annotation. In Proceedings of the 19th Linguistic Annotation Workshop (LAW-XIX-2025), pages 359–376, Vienna, Austria. Association for Computational Linguistics.

Ziwei Ji, Nayeon Lee, Rita Frieske, Tiezheng Yu, Dan Su, Yan Xu, Etsuko Ishii, Ye Jin Bang, Andrea Madotto, and Pascale Fung. 2023. Survey of hallucination in natural language generation. ACM Computing Surveys, 55(12):1–38.

Enkelejda Kasneci, Kathrin Seßler, Stefan Küchemann, Maria Bannert, Daryna Dementieva, Frank Fischer, Urs Gasser, Georg Groh, Stephan Günnemann, Eyke Hüllermeier, et al. 2023. ChatGPT for good? On opportunities and challenges of large language models for education. Learning and Individual Differences, 103:102274.

Salam Khalifa, Nizar Habash, Dana Abdulrahim, and Sara Hassan. 2016. A large scale corpus of Gulf Arabic. In Proceedings of the Tenth International Conference on Language Resources and Evaluation (LREC'16), pages 4282–4289, Portorož, Slovenia. European Language Resources Association (ELRA).

J. Peter Kincaid, Robert P. Fishburne, Richard L. Rogers, and Brad S. Chissom. 1975. Derivation of new readability formulas (Automated Readability Index, Fog Count and Flesch Reading Ease Formula) for Navy enlisted personnel. Research Branch Report 8-75, Naval Technical Training Command.

Rana Malhas, Watheq Mansour, and Tamer Elsayed. 2022. Qur'an QA 2022: Overview of the first shared task on question answering over the Holy Qur'an. In Proceedings of the 5th Workshop on Open-Source Arabic Corpora and Processing Tools with Shared Tasks on Qur'an QA and Fine-Grained Hate Speech Detection, pages 79–87, Marseille, France. European Language Resources Association.

Hamdy Mubarak, Rana Malhas, Watheq Mansour, Abubakr Mohamed, Mahmoud Fawzi, Majd Hawasly, Tamer Elsayed, Kareem Mohamed Darwish, and Walid Magdy. 2025. IslamicEval 2025: The first shared task of capturing LLMs hallucination in Islamic content. In Proceedings of The Third Arabic Natural Language Processing Conference: Shared Tasks, pages 480–493, Suzhou, China. Association for Computational Linguistics.

Ossama Obeid, Nasser Zalmout, Salam Khalifa, Dima Taji, Mai Oudah, Bashar Alhafni, Go Inoue, Fadhl Eryani, Alexander Erdmann, and Nizar Habash. 2020. CAMeL Tools: An open source Python toolkit for Arabic natural language processing. In Proceedings of the Twelfth Language Resources and Evaluation Conference, pages 7022–7032, Marseille, France. European Language Resources Association.